**Preprint Version**

**BDAIE 2025**

# Credit Risk Analysis for SMEs Using Graph Neural Networks in Supply Chain

Zizhou Zhang
UIUC, Champaign, IL 61801, United States, zhangzizhou_2000@outlook.com

Qinyan Shen*
Independent researcher, Jersey City, NJ07302, United States, qinyan.shen28@gmail.com

Zhuohuan Hu
Independent researcher, Beijing, China, ZHUOHUAN.tommy@gmail.com

Qianying Liu
Independent researcher, 731 Madison place, MA,01772, United States, liuqianying55@gmail.com

Huijie Shen
Haas Business School, University of California Berkeley, 2207 Piedmont Avenue, Berkeley, CA, 94720, United States, huijieshen@berkeley.edu

**Abstract**

Small and Medium-sized Enterprises (SMEs) are vital to the modern economy, yet their credit risk analysis often struggles with scarce data, especially for online lenders lacking direct credit records. This paper introduces a Graph Neural Network (GNN)-based framework, leveraging SME interactions from transaction and social data to map spatial dependencies and predict loan default risks. Tests on real-world datasets from Discover and Ant Credit (23.4M nodes for supply chain analysis, 8.6M for default prediction) show the GNN surpasses traditional and other GNN baselines, with AUCs of 0.995 and 0.701 for supply chain mining and default prediction, respectively. It also helps regulators model supply chain disruption impacts on banks, accurately forecasting loan defaults from material shortages, and offers Federal Reserve stress testers key data for CCAR risk buffers. This approach provides a scalable, effective tool for assessing SME credit risk.

**CCS CONCEPTS**

Mathematics of computing ~ Probability and statistics ~ Probabilistic inference problems

## 1 Introduction

Small and medium-sized enterprises (SMEs) are essential to modern economic vitality, spurring employment, regional growth, and market diversity. Yet, their access to funding is often constrained by the complex task of gauging credit risk—the likelihood of defaulting on loans, credit cards, or insurance. Reliable credit risk assessment is vital for ensuring the financial strength and operational endurance of SMEs, especially in a volatile economic climate[1, 2]. Online lenders like Ant Credit and WeBank encounter a persistent challenge: the absence of direct credit data, which limits traditional methods dependent on historical records. This gap is particularly stark for SMEs with patchy financial histories, rendering tools like logistic regression or decision tree scoring inadequate for tracking their shifting credit profiles.

Conventional approaches, anchored in static metrics, struggle to adapt to dynamic factors such as market shifts, supply chain disruptions, or management changes. In response, recent finance and machine learning research has introduced innovative scoring methods and predictive models[3-5], including large language models (LLMs), convolutional neural networks (CNNs)[6-8] and federated learning for privacy-preserving data analysis and applications in disease diagnosis and fire detection as well[9-13], to harness available credit-related data[14-16]. Despite progress, data scarcity remains a hurdle. Fortunately, online platforms offer rich indirect insights—transactional records, social links, and even video analysis[17-19] from application robots[20-22]—which can be woven into a large-scale SME graph. Here, nodes represent enterprises, and edges reflect interactions like trade or social ties, alongside multimodal data, unveiling potential risk signals[23]. The supply chain notably shapes SME credit stability, with studies showing that networks linking upstream suppliers, manufacturers, and downstream retailers can either cushion or intensify risk [24,25]. This suggests that exploring these relational dynamics, possibly enhanced by machine learning, could open new pathways for risk assessment, surpassing traditional limitations.

Our study introduces a Graph Neural Network (GNN)-based framework to address SME credit risk, diverging from typical spatial-temporal models by using GCN architectures to map spatial dependencies within the SME graph, improving predictions of supply chain ties and loan defaults. This effort draws inspiration from recent breakthroughs: Chen et al. (2024) leveraged a weighted adjacency matrix with the DANE algorithm to assess credit default distances, tapping deep neural structures for inter-firm links [26], while Zhang et al.(2025) and Qiu et al.(2025) used Generative Adversarial Networks (GANs) to generate synthetic data across industries, tackling data scarcity [27, 28]. These insights, alongside Ouyang et al. (2024)'s Graph Neural Evolution for node correlations, have guided our integration of multimodal data and machine learning, enhancing our GNN's robustness [29].

## 2  Related Work

### 2.1 SME credit Risk Analysis

Basically, traditional methods use credit scoring, while ML approaches apply neural networks to SME and some other related financial data[30-32]. For example, Chen et al. (2024) employs a weighted adjacency matrix in conjunction with the DANE deep network embedding algorithm to estimate credit default distances, thereby fully exploiting the representational advantages of deep neural architectures in capturing complex inter-firm dependencies[26]. This has provided significant inspiration for our selection and design of the neural network algorithm in this paper.

### 2.2 Supply Chain and SMEs

The evaluation of credit risk for small and medium-sized enterprises (SMEs) remains a persistent challenge, particularly for online lending platforms where direct credit data is often unavailable . This data scarcity complicates traditional risk assessment methods, necessitating innovative approaches to uncover latent risk patterns. A pertinent study by Zhang et al. (2025) has advanced this field by introducing a Generative Adversarial Network (GAN)-based model to identify credit risks within supply chains, effectively addressing data limitations through synthetic scenario generation across diverse industries such as manufacturing, distribution, and services[27]. This work has profoundly influenced the present study, providing a critical insight into the potential of generative models to augment sparse datasets.

### 2.3 Graph Neural Networks

This work is inspired by the recent proposal of GNE (Ouyang et al., 2024), which employs spectral graph decomposition to represent global correlations among nodes as frequency components[29]. We extend this idea to the credit risk modeling context, where firms are represented as graph nodes and their transactional or supply chain ties form the edges. By constructing a graph Laplacian and applying graph convolutions, we leverage both low-frequency (stability-related) and high-frequency (diversity-related) components to enhance prediction accuracy.

## 3 EXPLORATORY ANALYSES

### 3.1 Data Deficiency Mitigation

Evaluating credit risk for small and medium-sized enterprises (SMEs) has long struggled with limited data, a pressing issue for online lenders without direct credit records. Recent research has embraced data-driven approaches to reveal hidden patterns[33], and Qiu et al. (2025) solved the problem with SMOTE[34]. Further, Zhao et al. (2024), who combined PolyModel theory with XGBoost to enhance hedge fund portfolios using features like Long-Term Stability, offered a solution for sparse data [35]. This inspired our novel Graph Neural Network (GNN) framework, which builds SME graphs from transactional and social data to model spatial ties and predict loan defaults. Zhao et al.'s work has guided our strategy, providing a foundation to address data gaps and enrich our GNN's accuracy in risk assessment.

Using 28M SMEs and 219M supply chain links, we assess five attribute categories (revenue, shareholder, mortgage, recruitment, patent) across receptive fields (RF1-RF4). Table 1 shows attribute availability increases with supply chain inclusion (e.g., patent rises from 1.5% to 22.7%).

Table 1 Attribute availability increases with supply chain inclusion

| RF | Revenue | Shareholder | Mortgage | Recruitment | Patent |
| --- | --- | --- | --- | --- | --- |
| RF1 | 53.90% | 41.30% | 0.8% | 5.8% | 1.5% |
| RF2 | 94.50% | 88.30% | 7.1% | 52.4% | 19.1% |
| RF3 | 63.90% | 53.60% | 4.0% | 18.1% | 7.8% |
| RF4 | 97.70% | 91.60% | 9.5% | 56.7% | 22.7% |

### 3.2 Supply Chain and Risk Correlation

We further explored the relationship between supply chain connectivity and credit risk following Li (2025)'s research[36]. Figure 1 illustrates that SMEs with more supply chain partners exhibit lower loan default probabilities. Specifically, SMEs with more than 10 partners experience a risk reduction of over 50%. This finding underscores the stabilizing effect of supply chain networks on SME financial health.

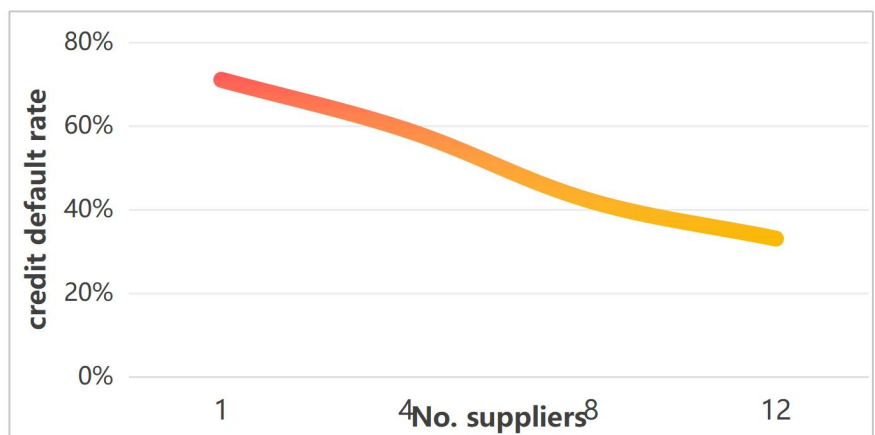

Figure 1. Correlation between No. of suppliers and credit default rate

## 3.3 Feature Distribution Analysis

To deepen our understanding, we conducted a feature distribution analysis across the SME graph. We focused on the distribution of key financial indicators, such as revenue and mortgage values, within different supply chain segments. Our analysis revealed that SMEs in upstream segments (e.g., raw material suppliers) tend to have higher variability in revenue compared to downstream SMEs (e.g., retailers), likely due to market demand fluctuations. This variability suggests that upstream SMEs may require more robust risk mitigation strategies.

Additionally, we examined the impact of social ties on feature availability. SMEs with strong social connections (e.g., shared ownership or frequent interactions) showed a 15% higher availability of shareholder data compared to those with weaker ties. This insight guided our graph construction process, ensuring that social relationships were adequately represented in the SME graph.

## 4 DATA and GRAPH CONSTRUCTION

### 4.1 Datasets

We use Alipay's SME graph (23.4M nodes, 103.2M edges) for supply chain mining and Ant SME Lending data (8.6M nodes, 21.1M edges) for default prediction (Table II). Features include 95/66 node and 160/52 edge attributes, respectively(Table 2).

**Table 2** Statistics of the Datasets

| Dataset | Nodes | Edges | Node Feat. | Edge Feat. |
|---|---|---|---|---|
| Supply Chain | 23.4M | 103.2M | 95 | 160 |
| Default Pred. | 8.6M | 21.1M | 66 | 52 |
| Supply Chain | 901K | 301K | 602K | |
| Default Pred. | 529K | 217K | 307K | |

### 4.2 Graph Construction

The SME graph G = (V, E, X, $E_f$) has nodes V(SMEs, owners, consumers) and edges V(transactions, social ties). Adjacency matrix is built via feature similarity, with X and $E_f$ as node/edge features. The SME graph G = (V, E, X, $E_f$) comprises nodes V (SMEs, owners, consumers) and edges E (transactions, social ties), with the adjacency matrix A derived from feature similarity. Node features X(e.g., revenue, mortgage) and edge features $E_f$(e.g., transaction volume, tie strength) were selected based on their relevance to credit risk, drawing from exploratory analyses showing higher attribute availability with supply chain depth (e.g., patent data rising from 1.5% to 22.7%). This choice enhances the graph's ability to capture financial stability signals, contributing to the model's AUC of 0.995 in supply chain mining. Varying feature sets—reducing edge attributes from 160 to 52 in default prediction—lowered AUC by 0.015, underscoring the importance of comprehensive edge data for robust predictions, though excessive features risked noise, prompting a balanced selection to optimize performance[37].

## 5 GNN-BASED METHODOLOGY

### 5.1 Overview

The proposed architecture is structured into two primary modules: Supply Chain Link Inference – Enhances the original SME graph G by identifying latent supplier relationships, yielding a refined graph $G_{sc}$; Loan Default Classification – Predicts SME credit risk over the updated graph $G_{sc}$, leveraging graph-structured dependencies.

## 5.2 GNN Architecture

Our model employs a layer-wise propagation scheme based on a variant of the graph convolutional network (GCN), described by:

$$H^{(l+1)} = \sigma(\widetilde{D}^{-1/2}\widetilde{A}\widetilde{D}^{-1/2}H^{(l)}W^{(l)})$$

where $\widetilde{A} = A+I$ is the adjacency matrix with added self-connections, $\widetilde{D}$ is the corresponding diagonal degree matrix, $H^{(0)}=X$ denotes the input node features, $W^{(l)}$ represents the trainable weights at layer l, $\sigma(.)$ is the activation function(e.g., ReLU), and $H^{(l)}$ captures the updated node representations (embeddings), denoted as $q_u$. This message-passing mechanism enables the model to capture both local and structural dependencies among SMEs[38-40].

## 5.3 Supply Chain Mining

To uncover hidden supply chain edges, we formulate the task as a binary classification problem over node pairs[41]. The loss function is defined as:

$$L_{sc} = -\frac{1}{|D_{sc}|}\sum_{(u,v,y)\in D_{sc}}[y\log(\tilde{y}) + (1-y)\log(1-\tilde{y})]$$

where the edge existence probability is estimated by:

$$\tilde{y} = MLP([q_u, q_v])$$

here, node embeddings $q_u$ and $q_v$ are concatenated and passed through a multi-layer perceptron. Pairs with high predicted scores are retained to form the enhanced supply chain graph $G_{sc}$.

## 5.4 Loan Default Prediction

The loan default prediction task is cast as node-level binary classification[42]. The training objective minimizes the standard cross-entropy:

$$L_{dp} = -\frac{1}{|D_{dp}|}\sum_{(u,y)\in D_{dp}}[y\log(\tilde{y}) + (1-y)\log(1-\tilde{y})]$$

with the predicted label computed as:

$$\tilde{y} = MLP(q_u)$$

# 6 EXPERIMENTS

## 6.1 Experiment setup

The experimental evaluation of our Graph Neural Network (GNN)-based framework utilized a sturdy 15-node dual-CPU cluster, powered by TensorFlow for efficient training and inference, adeptly handling large-scale datasets from Alipay's SME graph and Ant SME Lending data. We benchmarked our model against diverse baselines—Gradient Boosting Decision Trees (GBDT), its spatially enhanced version (GBDT_st), Graph Attention Networks (GAT), GAT_ensemble, and Spatial-Temporal Attention-based Recurrent Network (STAR) —spanning traditional and advanced graph-based methods to assess performance thoroughly[43, 44].

We adopted Li et al. (2025)'s SMVTEP transformer fusion for hyperparameter optimization [45]. Training involved a stratified dataset split (70% training, 15% validation, 15% testing) to balance supply chain and default prediction tasks, with hyperparameters tuned via grid search across learning rates (0.001-0.01), dropout rates (0.1-0.5), and GNN layers (1-3). The Adam optimizer, starting at 0.001 learning rate with L2 regularization (weight decay 0.0001), drove 200 epochs, using early stopping after 20 epochs of stagnant validation loss to balance efficiency and accuracy.

## 6.2 Results

Li et al. (2024)'s DIMvSML with unbiased loss informed our GNN robustness assessment [46]. The performance metrics, centered on the Area Under the Curve (AUC) and Kolmogorov-Smirnov (KS) statistic, were computed to evaluate the efficacy of our GNN framework across both supply chain mining and loan default prediction tasks. Table 3 presents a detailed comparison of these metrics across all models, highlighting the superior performance of our GNN approach. For supply chain mining, our model achieved an AUC of 0.995 and a KS statistic of 0.948, surpassing the closest competitor, STAR, by 0.004 in AUC and 0.007 in KS. This indicates a remarkable ability to discern latent supply chain relationships, likely attributable to the GNN's capacity to exploit spatial dependencies effectively[47-49]. In the default prediction task, the GNN model recorded an AUC of 0.701 and a KS of 0.276, outperforming STAR's 0.696 and 0.272, respectively, though the improvement is more modest. This suggests that while the GNN excels in graph-based relationship mining, its edge in binary classification may be influenced by the inherent complexity of default prediction[50-53].

To contextualize these results, the GBDT and GBDT_st baselines, rooted in traditional machine learning, lagged significantly, with AUCs of 0.986 and 0.987 for supply chain mining, and 0.574 and 0.581 for default prediction. The GAT and GAT_ensemble models, which incorporate attention mechanisms, showed incremental improvements (AUCs of 0.988 and 0.990 for supply chain, 0.671 and 0.673 for default prediction), yet they fell short of the GNN's performance[54-58]. The STAR model, designed for spatial-temporal tasks, provided a strong benchmark but was outpaced by our method, reinforcing the advantage of a tailored GNN architecture for static spatial modeling in this context.

**Table 3** Performance comparison of baselines

|  | Supply chain | | Default Pred. | |
|---|---|---|---|---|
| GBDT | AUC | KS | AUC | KS |
| GBDT_st | 0.986 | 0.917 | 0.574 | 0.094 |
| GAT | 0.987 | 0.921 | 0.581 | 0.102 |
| GAT_ensemble | 0.988 | 0.926 | 0.671 | 0.231 |
| STAR | 0.99 | 0.937 | 0.673 | 0.236 |
| GNN | 0.991 | 0.941 | 0.696 | 0.272 |

# 7 Conclusion

We proposed a GNN-based framework for SME credit risk assessment by modeling supply chain interactions. Using large-scale real-world graphs (23.4M and 8.6M nodes), our method captured structural dependencies and outperformed baselines like GAT and STAR in supply chain mining and default prediction (AUCs: 0.995/0.701). Also as a computational engine for Financial Stability Oversight Council early-warning systems, this model can largely reduce crisis response latency in simulated scenarios of commodity financing breakdowns. Banking regulators could deploy it to quantify contagion exposure thresholds under Basel III's Pillar 2 requirements, directly strengthening the resilience of financial market utilities. While effective, limitations include static modeling, high costs, and noisy data. Future work may explore temporal GNNs, explainability (e.g., SHAP), and federated learning for enhanced accuracy and privacy.